\documentclass[runningheads]{llncs}

\usepackage[english]{babel}
\usepackage{arxiv}

\usepackage{amsmath}
\usepackage{hyperref}
\usepackage{graphicx}
\usepackage{float}
\hypersetup{
     breaklinks=true,
     colorlinks=true,
     pdfusetitle=true,
     }

\title{The 2020 Presidential Election: Using Machine Learning to Understand Why Trump Lost}
\title{Fuzzy Forests For Feature Selection in High-Dimensional Survey Data:  An Application to the 2020 U.S. Presidential Election\\ \vspace{.25in}
\small{Paper prepared for presentation at The 3rd International Conference on Applied Machine Learning and Data Analytics, December 16-17 2021}}

\author{Sreemanti Dey\inst{1}\orcidID{0000-0003-2428-3538} \and
R. Michael Alvarez\inst{2}\orcidID{0000-0002-8113-4451} }

\author{}

\authorrunning{S. Dey and R.M. Alvarez}
\authorrunning{}

\titlerunning{Fuzzy Forests for Feature Selection in Survey Data}

\institute{}
\institute{California Institute of Technology, Pasadena, CA 91125\\
\email{sdey@caltech.edu} \\
\and
California Institute of Technology, Pasadena, CA 91125\\
\url{https://rmichaelalvarez.com}\\
\email{rma@caltech.edu}
}

\author{Sreemanti Dey and R. Michael Alvarez\thanks{Dey is an undergraduate research assistant at the California Institute of Technology and is the corresponding author (\email{sdey@caltech.edu}).  Alvarez is professor of political and computational social science at the California Institute of Technology (\email{rma@caltech.edu}).}}

\begin{document}
\maketitle

\begin{abstract}
    An increasingly common methodological issue in the field of social science is high-dimensional and highly correlated datasets that are unamenable to the traditional deductive framework of study. Analysis of candidate choice in the 2020 Presidential Election is one area in which this issue presents itself: in order to test the many theories explaining the outcome of the election, it is necessary to use data such as the 2020 Cooperative Election Study Common Content, with hundreds of highly correlated features. We present the Fuzzy Forests algorithm, a variant of the popular Random Forests ensemble method, as an efficient way to reduce the feature space in such cases with minimal bias, while also maintaining predictive performance on par with common algorithms like Random Forests and logit. Using Fuzzy Forests, we isolate the top correlates of candidate choice and find that partisan polarization was the strongest factor driving the 2020 presidential election.
\end{abstract}

\section{The Problem}

Social science research today often encounters a difficult methodological situation --- larger and larger datasets, which contain high-dimensional features, which are highly correlated \cite{grimmer2021machine}.  Quite literally, as in the application we discuss in our paper (the 2020 U.S Presidential election), to test the many different theories and potential explanations for why voters decided to remove then President Trump from office, researchers need to use methodologies that can quickly and efficiently reduce the feature space from hundreds of possible features to a smaller set that can then be the focus of further study.

In our paper we present a variant of the popular Random Forest, Fuzzy Forests, which we argue is well suited for exactly this type of applied machine learning problem \cite{conn2019fuzzy}.   Fuzzy Forests are ideal for feature selection in large and high-dimensional datasets, where the features are highly correlated.  Fuzzy Forests have seen use in medical science, health care, and in some social science applications \cite{toyama2019statistical}, \cite{jayasurya2016feature}, \cite{conn2019fuzzy}.  

In the next section we present the argument for the use of Fuzzy Forests, and then we provide details about how the Fuzzy Forests approach works.  We then discuss the details of the data and application, and thereafter present our results.  Substantively we find that  the 2020 election can be summarized as strongly driven by partisan polarization, with certain issue content most clearly illustrating the party divide. In particular, the results of our Fuzzy Forests show that the top correlates of candidate choice include party membership and past voting behavior, as well as opinions about filling the Supreme Court vacancy, racial relations, healthcare policy, and certain other well-known issues split on party lines. Age is the only demographic factor to appear within the top twenty variables, which is reflective of its role in the election.  We conclude with a discussion that recommends not only further use of Fuzzy Forests in social science applications, but more generally in applied machine learning settings.   

\section*{Applied Machine Learning in Social Science}

There has recently been much interest in applying machine learning to social science research questions. Ensemble methods such as Random Forest and Bayesian Additive Regression Trees have proven especially applicable \cite{grimmer2021machine}, \cite{montgomery2015informed}. Among the machine learning algorithms thus far used effectively, Fuzzy Forests specifically has also shown promise in finding the correlates of overspecified political questions such as candidate choice and voter turnout. We will show that Fuzzy Forests is also useful for studying candidate choice using large survey datasets.

In general, using machine learning in social science represents a methodological paradigm shift of sorts, from classical deductive studies to an inductive approach. In the past, most social science work has focused on hypothesis testing using established theories---this would normally involve hand-picking theory-related features from a dataset, building simple statistical models using a small and hand-picked set of features, and then testing hypotheses after the statistical model has been estimated using the available dataset.  The problem of candidate choice is not amenable to such a study. First, there is an overabundance of theories that could explain candidate choice, and they are not mutually exclusive by any means. There is, therefore, difficulty in choosing survey features to include---this inability to choose specific features to study is another deviation from the deductive framework. Second, there is a stricter validation requirement for candidate choice results, due to the larger problem space.

We argue that an inductive study is better suited to this problem, and machine learning provides a suitable methodology. We choose a mix of theories about candidate choice to provide context for our findings, but we do not constrain our data features to fitting the hypotheses. Instead, from there, we follow the framework for using machine learning in social science, with discovery, measurement, and inference stages \cite{grimmer2021machine}. Notably, these stages are essentially built-in steps in the Fuzzy Forests model (which as we discuss in detail in the next section is an extension of Random Forests \cite{breiman2001random}). Our discovery stage is the module formation stage of the algorithm, essentially clustering the features of the data themselves. The measurement stage in Fuzzy Forests has two main components: unsupervised dimensionality reduction, and supervised learning using the remaining features. Finally, our inference stage is mainly prediction--we use k-fold cross validation and ROC curves to observe the efficacy of our model.

Clearly, Fuzzy Forests is uniquely suited to our methodological needs, and is a regression-tree machine learning approach that is useful for larger survey datasets \cite{kern2019tree}. As for substantive results, our approach to using this algorithm for candidate choice analysis can be classified as a fictitious prediction problem \cite{grimmer2021machine}. We train our model to predict candidate choice according to the framework laid out above, but our goal is actually to observe what features it is using to make the predictions, to draw conclusions about what factors were most strongly related to candidate choice in 2020.

\section{Why Fuzzy Forests?}

The Fuzzy Forests algorithm is an extension of Random Forests that is adapted to working with high-dimensional, correlated datasets. Specifically, its function is to identify some number of important features from such a dataset, from which a predictive model can then be constructed. Fuzzy Forests is designed to overcome the bias in feature-selection for these kinds of datasets that is present in other algorithms such as LASSO, SCAD, or Random Forests \cite{conn2019fuzzy}. The Fuzzy Forests method has been used in a number of biomedical and social science research studies \cite{toyama2019statistical}, \cite{jayasurya2016feature}.  There are three main stages to Fuzzy Forests: correlated features are first grouped together into modules, RFE-RF is used to screen features within modules, and finally, some of the remaining features are selected as the most important correlates through an aggregate RFE-RF.

While Fuzzy Forests allows module formation by any method, the default is by Weighted Gene Correlation Network Analysis (WGCNA) \cite{zhang2005general}. WGCNA first calculates the connection strength between pairs of features by using a similarity function, usually closely related to the Pearson correlation, and raising it to a power $\beta$. These values are aggregated, and an adjacency function is applied in order to make an adjacency matrix. Then, the topological overlap is calculated between each pair of features by checking whether both features are strongly connected to the same group of features. Once the topological overlaps are calculated, hierarchical clustering is used to separate the features into modules. The goal is for there to be high correlation within a module and weak correlation between modules. 

Once the modules have been constructed, the screening step begins. Within each module, Recursive Feature Elimination Random Forests (RFE-RF) are used to prune features. A Random Forest is fit within a module, and some proportion of features, for example, 25\%, with the lowest VIMs are dropped. This is done until the stopping criteria is reached, to be defined by the user. 

Lastly, the selection step is carried out by fitting one RFE-RF to all the remaining features after screening. Thus, the user is left with the desired number of important features, from which a Random Forest can be constructed. Predictive analysis is carried out using this Random Forest.

\section{Application:  The 2020 U.S. Presidential Election}

\subsubsection{Many Explanations, Many Features}

There are a myriad of factors that might account for Trump's loss in November 2020.  One set of factors are demographic, including age, gender, education, and income. The 2020 election saw increased turnout by younger voters--a demographic that favored Biden by a large margin. Another change was the decreased gender gap between each party's white voters: a greater proportion of white women voted for Trump and a greater proportion of white men voted for Biden in 2020 than for Trump and Clinton in 2016, respectively \cite{igielnik_keeter_hartig_2021}. Biden also made gains with white voters without college degrees.

Along with demographic shifts, there were many hot-button political issues that were highly salient in 2020. These included racial tensions from the recent Black Lives Matter protests, healthcare policy (aka ``Obamacare''), the Supreme Court vacancy left by the passing of Justice Ruth Bader Ginsberg, social issues like abortion and immigration reform, and the ``Trump Effect".\footnote{A colloquialism for the divisive and racially inflammatory rhetoric used by Trump \cite{newman2021trump}.} And of course, American voters are highly polarized along partisan and ideological lines, meaning that these political identities no doubt played some role in voter decisions in this election.

One further set of factors that might have played important roles in voter decision making in the 2020 presidential election regarded various aspects of Trump's performance in office since 2017.  These include his rhetoric and actions regarding race relations and immigration, and with respect to nationalism and foreign policy \cite{reny2019vote}, \cite{sides20172016}.  As the COVID-19 pandemic hit the United States in early 2020, Trump's response to the pandemic ranged included initiation of Operation Warp Speed to quickly develop vaccines, his embrace of various therapies to avoid or treat those infected with the virus (for example, monoclonal antibodies), his unwillingness to endorse many of the measures to prevent viral spread recommended by public health professionals, and his own COVID-19 infection, all served to politicize the pandemic.  Finally, the pandemic also upended the American economy in 2020, with many people losing jobs or suffering financial loss, though state and federal economic policies may have served to mitigate these economic repercussions of the pandemic.

Based on these arguments and on past research, we formulate five primary hypotheses that we wish to test.
\begin{enumerate}
\item{Those who show Republican or conservative leanings, either in self-identification or on symbolic issues, will vote for Trump, while those that lean Democratic or liberal  will vote Biden. }

\item{Opinions on racism, racial policies, opinions on sexism, or any other opinions regarding specific demographics will be closely tied to candidate choice.}

\item{Policy opinions will be present within the top correlates of candidate choice, showing that issue voting was a driver of candidate choice in 2020.}

\item{Opinions and circumstances relating to COVID-19 will be top correlates of candidate choice, showing that pandemic blame attribution strongly affected voting.}

\item{Recent changes in personal financial, opinions on the national economy, or both will be correlated with candidate choice, evidencing retrospective economic voting.}
\end{enumerate}

Of course, we have just outlined a wide number of potential explanatory factors that might play important roles in Trump's 2020 electoral loss, and we will discuss these in the context of voting behavior theory below.  But with such a wide array of potential explanations, each one of which can be operationalized using many different survey questions, the high dimensionality of the feature space makes use of the traditional methods for studying voting behavior with individual-level survey data problematic.  This is exactly the type of research problem that Fuzzy Forests can solve.

Thus, we apply machine learning, specifically, the novel Fuzzy Forests algorithm, in our paper. To our knowledge, we are among the first to apply machine learning to the topic of candidate choice, and certainly the first to apply Fuzzy Forests to this topic. Fuzzy Forests is an ensemble method well-suited for candidate choice analysis: it is designed to find the most important features of a correlated, high-dimensional dataset, which we expect to see in the survey data that we utilize in this paper. By using Fuzzy Forests, and machine learning in general, we frame our study as an inductive one, granting necessary flexibility to both our hypotheses and results.

\subsection{Data}

The data for this study comes from a preliminary release of the 2020 Cooperative Election Study Common Content (CES) dataset (previously the Cooperative Congressional Election Study, CCES). The CES dataset uses stratified sampling and in total, contains 61,000 survey responses. Of these, we only worked with respondents who said that they voted for either Joe Biden or Donald Trump in the 2020 presidential elections, limiting our sample size to 43,890. Since the data used was the pre-release, there is no voter validation, so we assume that respondents who say they voted did indeed vote.

Out of all the questions in both the pre-election and post-election survey, we chose 124 to include in our model. Key demographic questions regarding gender, race, age, ethnicity, religion, etc. were included, as well as several ideological questions and opinions about current issues, such as US racial relations, abortion, healthcare policy, and others. We aimed to encompass as wide of a political scope as possible with our feature choices.

Within our resulting dataset, we observed a missingness of about $1.51\%$, and used Predictive Mean Matching (PMM) hotdeck imputation to fill in the missing information. PMM chooses a set of candidate donor observations for a particular missing observation, and one of the donors is chosen for the imputation. This assumes that missing data follows the same distribution as known data, and ensures that imputed values are within the range of observed values. Given the relatively low incidence of missing data in our sample, we argue that this is a reasonable assumption.

Finally, the data was one-hot encoded, giving 443 variables that show a high level of correlation with each other, a very suitable dataset for Fuzzy Forests.

\begin{figure}[tbh]
\centering
\includegraphics[scale=0.6]{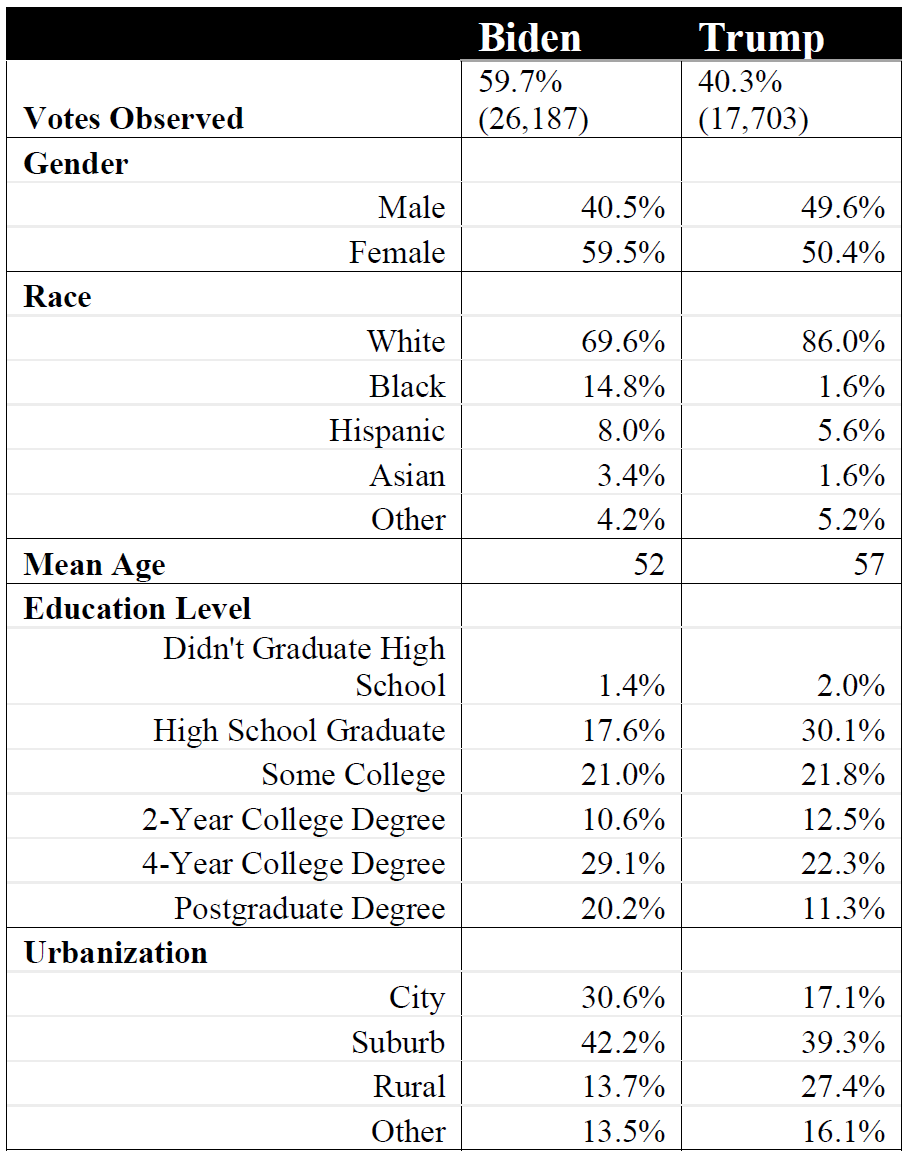}
\caption{\label{fig:dem_table}A breakdown of candidate choice in our dataset based on selected demographic variables.}
\end{figure}

From Figure 1, it can be seen that most of the demographics based on candidate choice in our dataset match the literature about each candidate's voter base. The proportion of females happens to be unusually large, especially among those who voted Biden, but the relative proportions of other variables are approximately as expected. We observe that Biden voters are more diverse, a little bit younger, have a greater proportion of high education levels, and are more urbanized than Trump voters. Although the data is basically predictable, the dataset also includes a weighting variable that accounts for bias present in the CES data, to further reduce disparities between the survey data and real voter behavior.

\begin{figure}[tbh]
\centering
\includegraphics[scale=0.35]{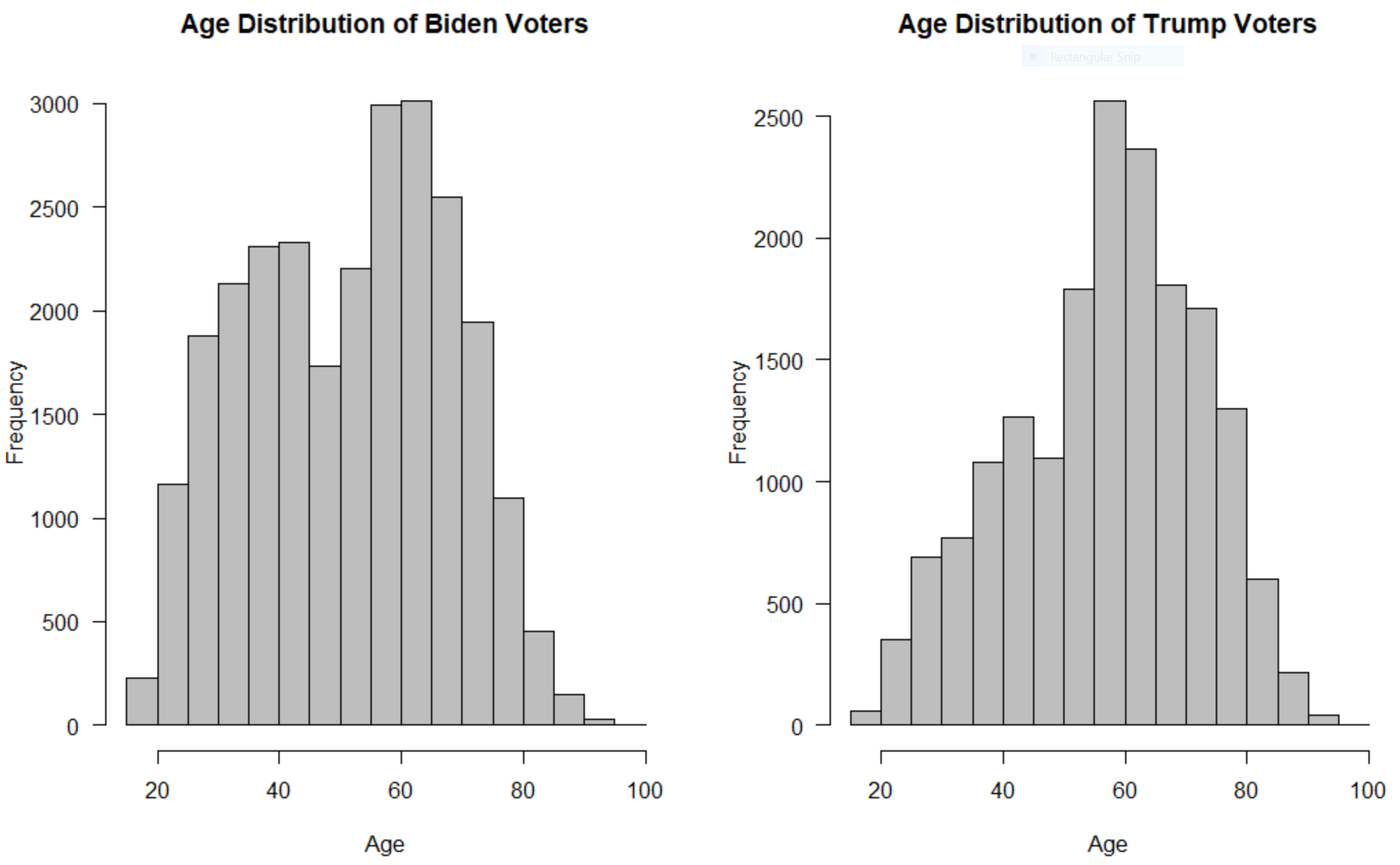}
\caption{\label{fig:dem_table}Age distributions of Biden and Trump voters.}
\vspace{.1in}
\parbox{5in}{Note: Biden voters have a bimodal distribution, while Trump voters do not. Both have a peak in the 50-70 range.}
\end{figure}

It is worth considering the age of the electorate more closely---although the mean ages of Biden and Trump voters are not drastically different, the respective distributions of ages demonstrate that younger people tend to favor Biden. This is consistent with demographic shifts in the electorate that were observed for the election at large. Considering this distribution difference as well as its implications about partisanship-fueled voter turnout, we expect age to be an especially important demographic feature in Fuzzy Forests.

\begin{figure}[tbh]
\centering
\includegraphics[scale=0.5]{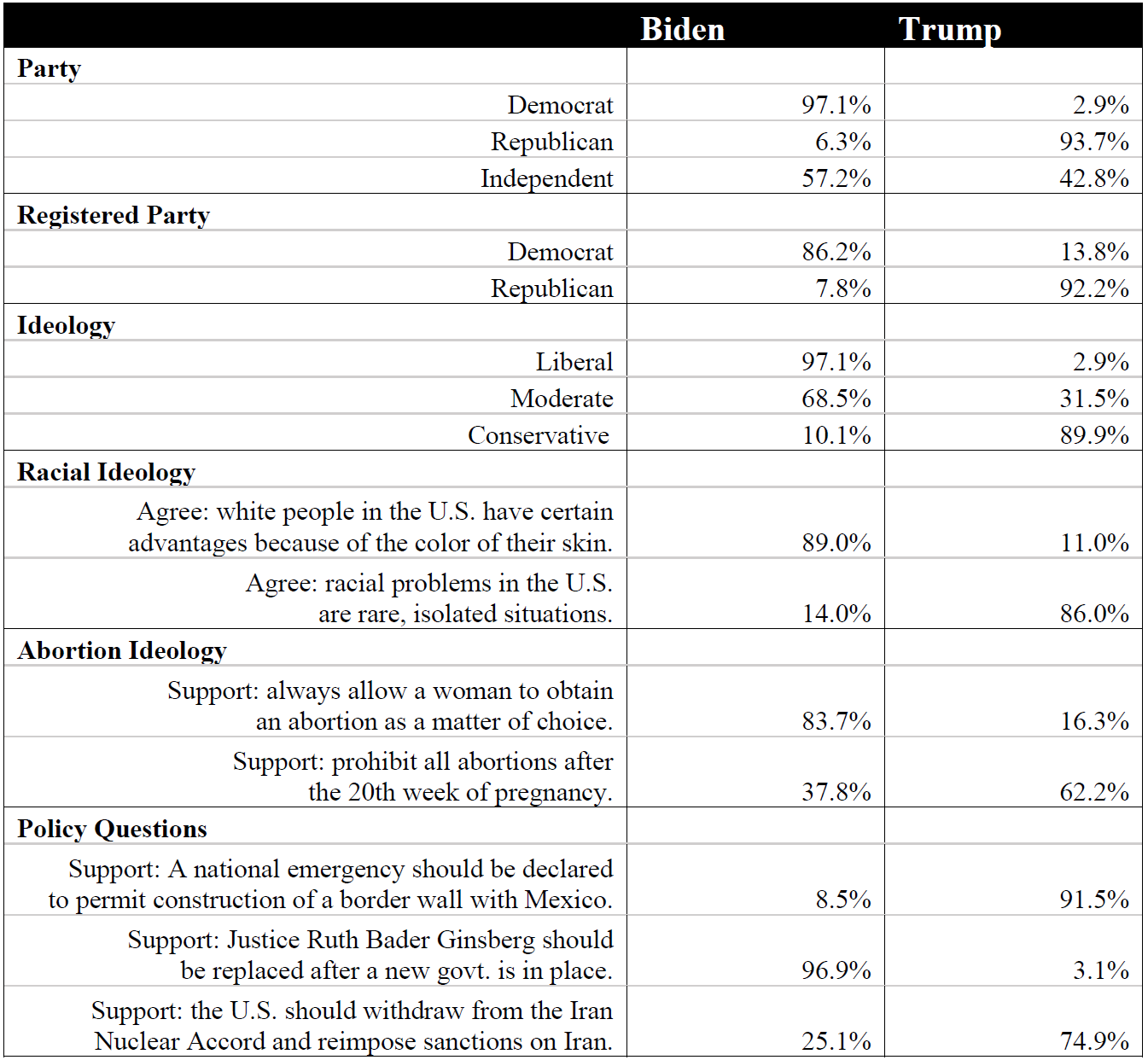}
\caption{\label{fig:dem_table}Candidate choice based on selected political variables. }
\end{figure}

Figure 3 shows that ideological questions are much better correlates of candidate choice. Party, ideology, and the first two policy questions are especially divisive, so we expect them to have high variable importance in the results of the Fuzzy Forests algorithm as well. In fact, we expect ideological questions to make up the bulk of important variables, based on the below breakdown.

\section{The Fuzzy Forests Result:  Why Trump Lost The 2020 Election}

The Fuzzy Forests model yields some interesting results.\footnote{Supplemental materials along with complete code needed for replication are available at \url{https://github.com/sreemanti-dey/fuzzy_forests_2020_election}. Supplemental materials include an additional ROC/AUC comparison, Fuzzy Forests hyperparameter choices and their justifications, and complete model feature and module membership information.}  The first of these are the modules themselves resulting from WGCNA, and the second is the list of most important variables from the dataset. Our conclusions will be drawn from the latter of these, as the modules are not designed to provide substantive results.

As part of our model validation, we take ROCs and AUCs of three different models, shown in Figure 4. We observe that the twenty variables returned by Fuzzy Forests have very nearly the same predictive power as all 124 variables in our dataset, as Figure 4 shows. This demonstrates that our Fuzzy Forest model is robust, and measures up to more traditional models. 

\begin{figure}[tbh]
\centering
\includegraphics[scale=0.4]{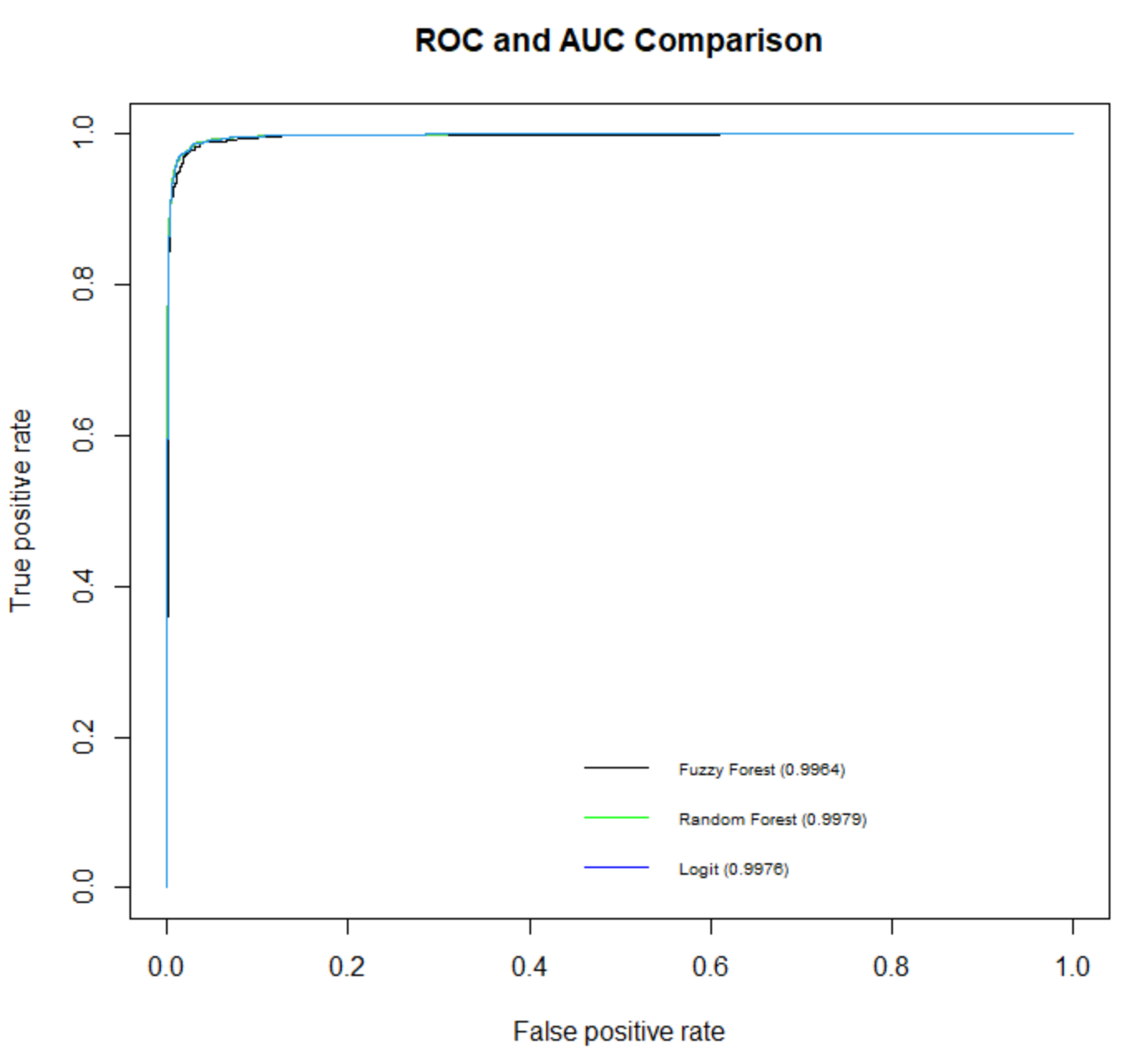}
\caption{\label{fig:dem_table}ROC curves and AUCs of the best performing models from each cross validation.}
\vspace{.1in}
\parbox{5in}{Note:
The Fuzzy Forests curve corresponds to a Random Forest trained on the top twenty variables resulting from Fuzzy Forests, while the Random Forests and logit were both trained on all features of the dataset.}
\end{figure}

\subsection{Module Formation}

WGCNA analysis on the dataset yielded the module membership shown in Figure 5. Due to the nature of one-hot encoding, variables and their opposites overwhelmingly appeared in the same module (i.e. ``yes" and ``no" responses to the same question both appeared in the same module). To account for this, the minimum module size was set to 5, so that it became likelier for at least two distinct questions to appear in a module. It also prevents the formation of an excessive number of modules more effectively and organically than simply limiting module formation.

The resulting modules group certain categories of questions together, especially ones that were related in the original survey; for example, the blue module contains political involvement questions, brown includes opinions on troop deployment, etc. An exception is the grey module, which serves to group variables that are not highly correlated to others. It is interesting that most ``neutral" answers (e.g. ``not sure" or ``neither agree nor disagree") overwhelmingly ended up in the grey module--the features of interest to the model, then, are ones that illustrate at least moderately strong opinions. Some of the colors are more nebulous in meaning, and it is not recommended to attempt to draw conclusions about voter blocs from module membership for this reason. For example, it is difficult to ascertain the ``genre" of the black module.

The most notable of these is turquoise, which is a large module containing highly partisan questions. Since module formation depends on the detection of similar answers, there is clearly an extremely strong sense of ideological divide in the survey responses. We expect, based on this module formation and our data crosstabs, that the turquoise model will be well-represented among our top variables.

\begin{figure}
\centering
\includegraphics[scale=0.6]{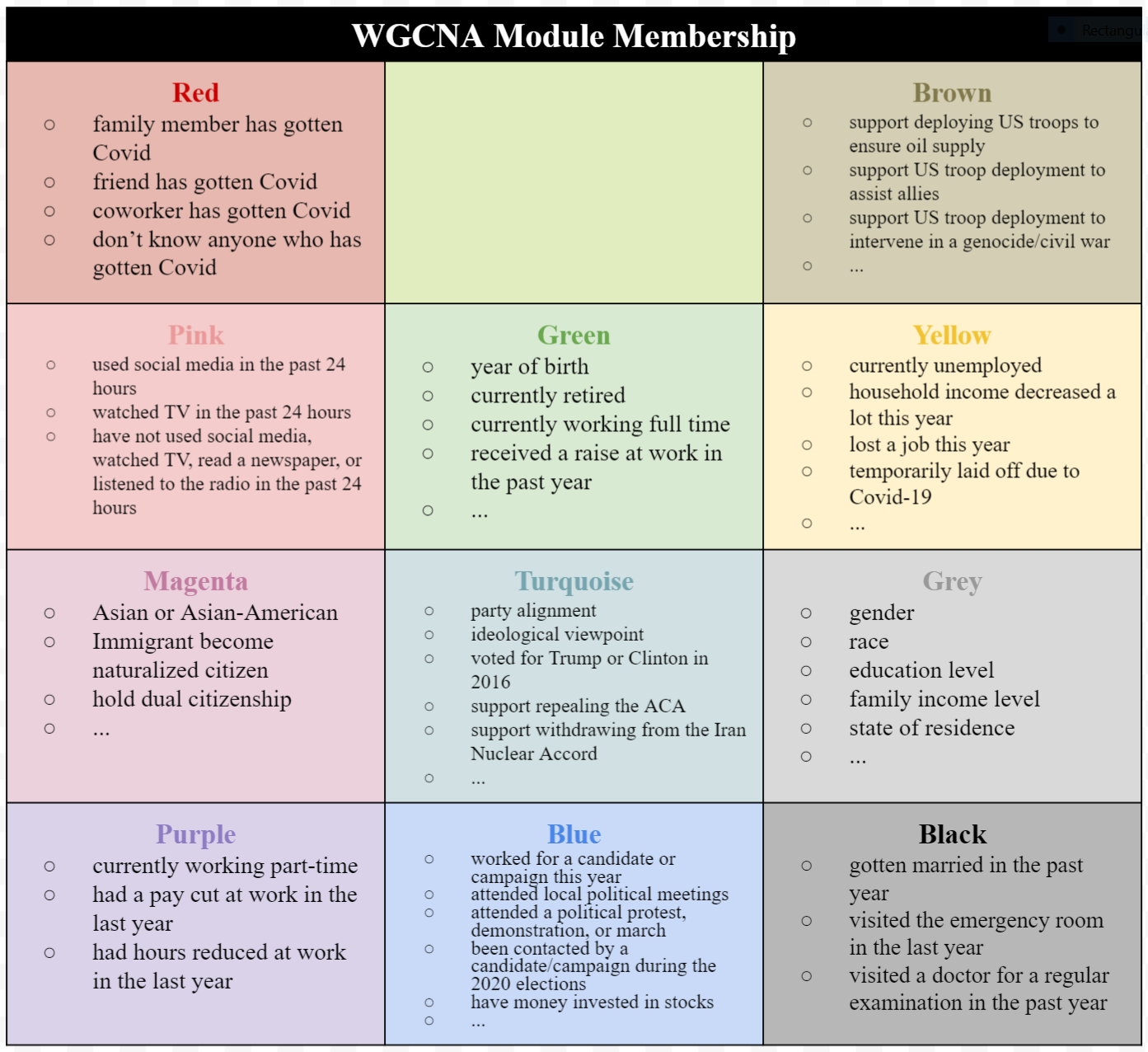}
\caption{\label{fig:dem_table}Module membership of certain model features calculated by WGCNA.}
\vspace{.1in}
\parbox{5in}{Note:
See Appendix C in Supplemental Materials for complete module membership information.}
\end{figure}

The modules are then passed into Fuzzy Forests, which uses recursive feature elimination to trim down each module by 25\% in each round. Again, pairs of responses to a question are chosen to drive each module.

\subsection{Variable Importance}

Finally, after three rounds of feature elimination within modules, feature elimination is conducted on the model as a whole, leaving a user-specified number of features remaining. We chose to keep 20 out of the original 443 features, and Figure 6 shows what these features represent. 19 out of these 20 features come from the turquoise module, which solidifies the notion that partisan opinions are very good correlates of candidate choice, more so than almost any demographic factor.

\begin{figure}[tbh]
\centering
\includegraphics[scale=0.5]{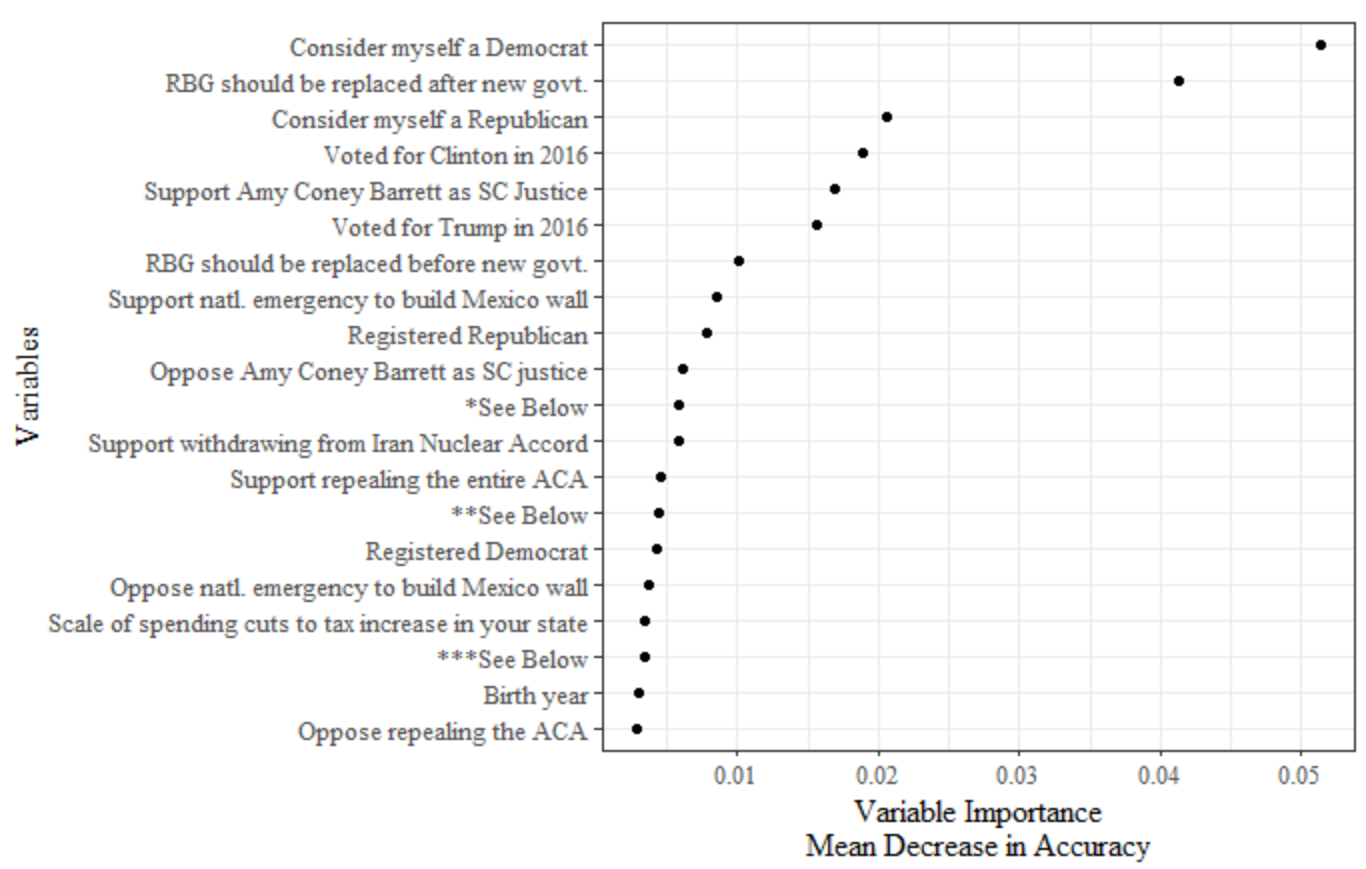}
\caption{\label{fig:dem_table}The top 20 most important features selected by the Fuzzy Forests model.}
\vspace{.1in}
\parbox{5in}{Note:*Strongly agree: white people in the U.S. have certain advantages because of the color of their skin. **Strongly agree: generations of slavery and discrimination have created conditions that make it difficult for Blacks to work their way out of the lower class. ***Strongly disagree: generations of slavery and discrimination have created conditions that make it difficult for Blacks to work their way out of the lower class.}
\end{figure}

The top two correlates of vote choice, in all runs of Fuzzy Forests, are ``Consider myself a Democrat" and ``Justice Ruth Bader Ginsberg should be replaced after a new govt. is in place in January 2021." This is somewhat surprising--a political opinion question consistently has a higher VIM than the counterpart to the most important feature, which would be ``Consider myself a Republican." In fact, among the six features with VIMS greater than 0.01, it is notable that two of them are related to the vacancy in the Supreme Court. The others are straightforward; it is expected that voting behavior will conform to party affiliation, and that party choice will remain consistent across multiple elections.

It should be noted, however, that the content of the top features is more important than their specific VIMs--although Fuzzy Forests reduces bias in VIMs from Random Forest, the final RFE-RF step still has the possibility of introducing correlation bias \cite{conn2019fuzzy}.

We gauge the robustness of the top variables by comparing their predictive power against the full dataset. This is done by checking the ROC curves of the 20-feature Random Forest resulting from Fuzzy Forests against a Random Forest trained on all the features, as well as a logit model.

Again, Figure 4 shows that the final Random Forest generated by Fuzzy Forests is on par with both the full Random Forest and logit. In fact, all three of the models have nearly perfect predictive power.

\subsection{Discussion}

The ideological features in the top 20 variables illustrate some of the issues that can be considered the most polarizing between Biden and Trump voters. In light of the overwhelming representation of ideology as a correlate of candidate choice, it is interesting that age happens to be the only demographic variable that makes it to the top twenty. This is particularly surprising because although it is known that there is a tendency for young voters to be left-leaning, as Figure 2 shows, age still appears among far more clear-cut and polarizing differences.

Our results strongly match our first hypothesis about candidate choice, that the 2020 election was highly partisan. Party is undeniably a big factor, as Democrat or Republican identification as well as party registration and previous voting are all among the most important variables, consistent with past research \cite{campbell1980american}, \cite{johnston2006party}. Symbolic issues are certainly present: the Supreme Court, border wall, and Iran Nuclear Accord questions are highly politicized even though they would not have a conceivable effect on most of the population \cite{sears1980self}. However, many self-interest policies are also present among the top twenty, such as the ACA question and the tax-spending cuts continuum, but these are highly politicized within party lines as well. 

The second hypothesis regarding candidate choice, identity politics, is also visible in the output of Fuzzy Forests. Racial questions have a large role, as three of the top twenty features are direct opinions about US racial relations, and the border wall is a highly racialized issue--implying that, as in 2016, racial opinions played an important role in the 2020 election \cite{algara2019distorting}, \cite{schaffner2016explaining}. None of the top twenty features were directly related to sex and gender, however, which matches the decreased gender gap among white voters observed in the 2020 election compared to 2016 \cite{bracic2019sexism}.

Somewhat surprisingly, issue voting, as per our third hypothesis, is also strongly present in the top twenty variables (more strongly than racial questions, in fact). Some of the policy questions are indeed easy issues, by the metrics of Carmines and Stimson, such as withdrawing from the Iran Nuclear Agreement and opinions on the Mexico border wall \cite{carmines1980two}. However, ``hard issues" also make up a portion of policy in the top twenty--for example, the Supreme Court replacement questions do not qualify as easy issues by any parameter, yet they consistently have very high variable importance. The state spending cut question is also a hard issue. Regardless of easy or hard, issues make up half of the most important variables--but, rather than attempting to claim a level of sophistication in candidate choice selection by voters that has never been seen before, it is more reasonable to claim that these issues have been so politicized that voters may simply base their opinions on what their party says \footnote{Johnston succinctly phrases this idea, put forth by Campbell et al., as ``such policy orientations as do appear [among voters] are more effects than causes of partisanship \cite{johnston2006party}."}, which brings us back to partisanship.

Our fourth hypothesis, COVID-19 blame attribution, was not supported by our top variables. We cannot say conclusively that pandemic policy was not related to voting, because the CES dataset did not include questions regarding opinions about specific COVID policies. But, with the exception of witnessing or going through COVID oneself, which is part of the data, the likeliest other source of pandemic frustration would have been economic--yet, neither direct COVID experience nor economic policies are strongly present in the top variables.

By extension, our fifth hypothesis, retrospective economic voting, is also not supported. As mentioned, economic policies are not among the top variables, and neither are opinions on the national economic conditions or personal financial situations.

From our set of hypotheses, we observe that the ones most strongly supported are all closely related to partisanship, along with our first hypothesis, partisanship itself.

\section{Conclusion}

The 2020 election saw massive turnout and a very close victory margin for Biden. Using Fuzzy Forests, we find that the factors most closely associated with candidate choice in the election were extremely partisan. These include party identification, past candidate choices, the Supreme Court vacancy, opinions on race, healthcare policy, and other widely politicized issues. Age was also found to be an important correlate.

Our results match a good amount of previous literature regarding vote choice, especially works that identify political parties as cohesive structures inspiring group loyalty. We also observe a strong relationship between racial attitudes and candidate choice, which seems to be a continued trend from the 2016 election. 

We also show that machine learning methods are useful for answering political science questions with a large observation scope. In particular, Fuzzy Forests has proven parsimonious and accurate for isolating factors best correlated with candidate choice in the 2020 election while simultaneously offering insight into networks in the data with WGCNA. 

Further research should aim to conclude a causal relationship between the important variables found by Fuzzy Forests and candidate choice, to check if the associations found match voter thought processes. Another direction for research is to apply Fuzzy Forests to other aspects of political science, the most closely related being voter turnout in the 2020 election.

Machine learning methods have proven to be useful in a wide variety of applications, and political science is no different.The future of candidate choice research will involve using machine learning methods such as Fuzzy Forests to quickly and accurately assess voter behavior.


\bibliographystyle{splncs04.bst}
\bibliography{mybibliography}

\end{document}